\documentclass[conference]{IEEEtran}
\usepackage[font=footnotesize,labelfont=bf]{caption}
\IEEEoverridecommandlockouts
\usepackage{cite}
\usepackage{amsmath,amssymb,amsfonts}
\usepackage{graphicx}
\usepackage{subcaption}
\usepackage{multirow}
\usepackage{makecell}
\usepackage{algorithm}
\usepackage{algpseudocode}
\usepackage[dvipsnames]{xcolor}
\usepackage{textcomp}
\usepackage{xcolor}
\usepackage{booktabs}
\usepackage{url}
\def\BibTeX{{\rm B\kern-.05em{\sc i\kern-.025em b}\kern-.08em
    T\kern-.1667em\lower.7ex\hbox{E}\kern-.125emX}}

\usepackage[hidelinks]{hyperref}
\begin{document}

\title{Bridging the Language Gap: Enhancing Multilingual Prompt-Based Code Generation in LLMs via Zero-Shot Cross-Lingual Transfer}

\author{
Mingda Li\textsuperscript{1,2},
Abhijit Mishra\textsuperscript{2}\thanks{Corresponding author: abhijitmishra@utexas.edu},
Utkarsh Mujumdar\textsuperscript{2}\\
\textsuperscript{1}Department of Statistics and Data Science, Yale University, New Haven, USA\\
\textsuperscript{2}School of Information, University of Texas at Austin, Austin, USA\\
Email: \{mingdali, abhijitmishra, utkarsh.mujumdar\}@utexas.edu
}

\maketitle

\begin{abstract}
The use of Large Language Models (LLMs) for program code generation has gained substantial attention, but their biases and limitations with non-English prompts challenge global inclusivity. This paper investigates the complexities of multilingual prompt-based code generation. Our evaluations of LLMs, including \textsc{CodeLLaMa} and \textsc{CodeGemma}, reveal significant disparities in code quality for non-English prompts; we also demonstrate the inadequacy of simple approaches like prompt translation, bootstrapped data augmentation, and fine-tuning. To address this, we propose a zero-shot cross-lingual approach using a neural projection technique, integrating a cross-lingual encoder like LASER to map multilingual embeddings from it into the LLM's token space. This method requires training only on English data and scales effectively to other languages. Results on a translated and quality-checked \emph{MBPP} dataset show substantial improvements in code quality. This research promotes a more inclusive code generation landscape by empowering LLMs with multilingual capabilities to support the diverse linguistic spectrum in programming.
\end{abstract}

\begin{IEEEkeywords}
Multilingual Artificial Intelligence, Code Generation, Machine Translation, Transformer
\end{IEEEkeywords}

\section{Introduction}

\begin{figure*}[t]
  \centering
  \includegraphics[width=\textwidth]{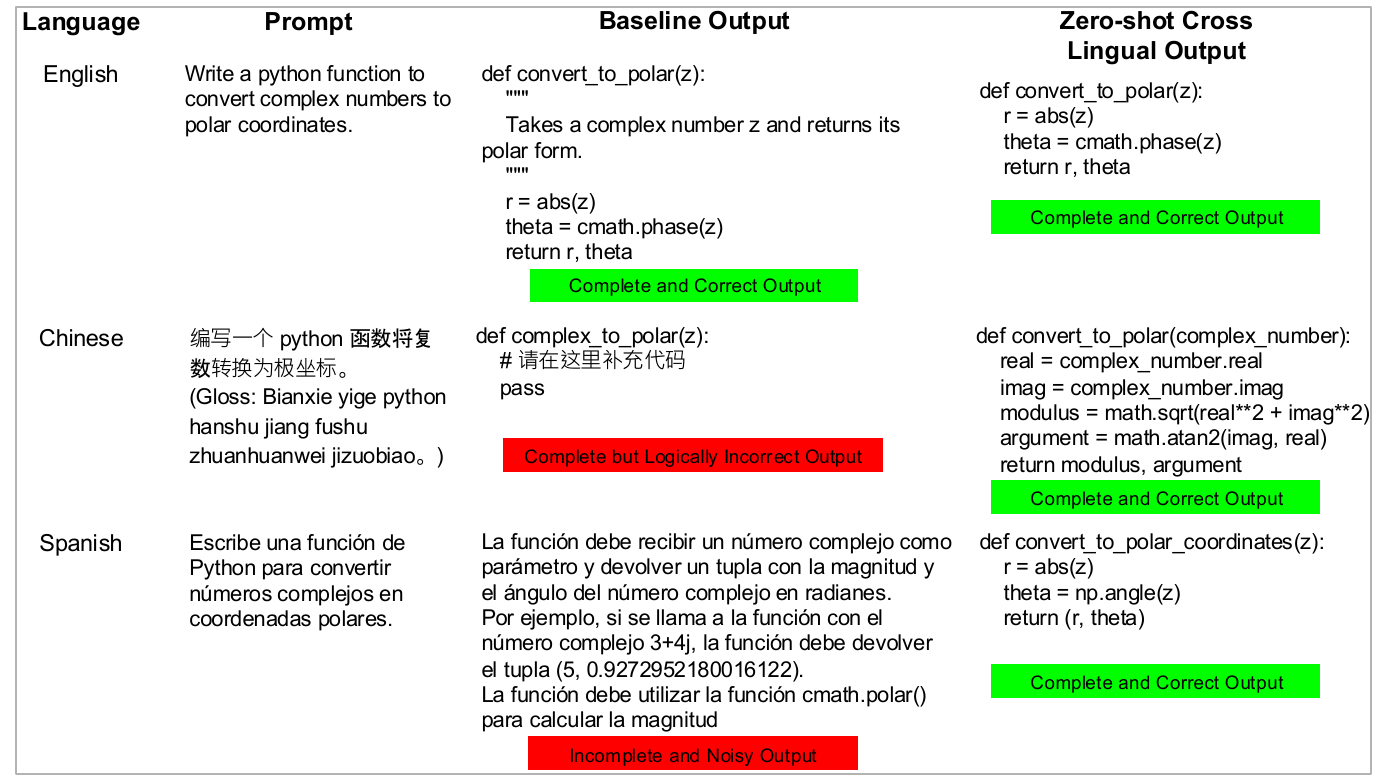}
  \caption{Disparity in output code generated by \emph{CodeLLaMa-Instruct} model\cite{roziere2023code} with 7B parameters for the same problem statement given in multiple languages}
  \label{fig:example}
\end{figure*}

The use of Large Language Models (LLMs) for code generation, such as generating Python programs from problem specifications, has gained substantial interest due to their effectiveness in handling complex language tasks \cite{zhao2023survey,gao2023pal,austin2021program}. This capability has led to the development of innovative applications like GitHub Copilot \cite{yetistiren2022assessing} and specialized LLMs such as CodeLLaMa \cite{roziere2023code}, underscoring the growing importance of this field. Although LLMs are globally prevalent and proficient in processing multilingual inputs, they often exhibit biases against non-English prompts \cite{talat-etal-2022-reap,choudhury2021linguistically}, which is particularly evident in code generation. Figure \ref{fig:example} under \textit{Baseline Output} illustrates how the quality of generated code diminishes as prompts shift from English to other languages, a disparity linked to the availability of data used in LLM training and fine-tuning. Ensuring LLMs deliver consistent quality across languages  is crucial for fostering fair and inclusive code generation, especially as the global programming community is increasingly composed of non-English speakers. Data from coding platforms like The Competitive Programming Hall of Fame\footnote{https://cphof.org/countries} highlight the skew towards non-English-speaking regions. As the global population of coders grows, addressing these biases in LLMs is essential to promoting an equitable and accessible environment for developers worldwide.

This paper first tries to bridge the language gap in multilingual prompt-based code generation by enhancing LLMs through self-supervised fine-tuning. To begin with, we evaluate the performance of LLMs such as \textsc{CodeLLaMa} and \textsc{GPT-4} on English and five non-English languages, by translating and quality-checking a sanitized version of the MBPP dataset \cite{austin2021program}. Significant disparities in code quality across languages are observed, even when using the same problem statements. Inspired by previous work on improving LLM performance in multilingual tasks with Chain-of-Thought (CoT) \cite{shi2022languagemodelsmultilingualchainofthought, qin2023crosslingualpromptingimprovingzeroshot}, and using bootstrapping for multilingual data generation \cite{awasthi2022bootstrapping}, we form strong baselines for code generation with CoT and fine-tuning on bootstrapped multilingual data. However, these approaches showed only marginal and inconsistent improvements.

To address data sparsity and limited multilingual exposure, we propose a novel approach: (a) using a pre-trained multilingual encoder like LASER \cite{artetxe2019massively} to encode multilingual inputs into a joint vector space; (b) projecting these embeddings into the LLM’s input space and aligning them through training solely on English data; (c) employing this LASER$\rightarrow$LLM pipeline at inference for zero-shot cross-lingual processing of non-English inputs. This method familiarizes models with multiple languages without needing additional external training data. Our evaluation shows improved code quality and reduced syntax and logical errors, as illustrated in Figure \ref{fig:example} under \textit{Zero-shot Cross-Lingual Inference}. Our approach integrates seamlessly as a minor training step that does not require any expensive pretraining or fine-tuning, thus offering a promising way to enhance LLMs' multilingual capabilities in code generation.

The contributions of the paper are as follows:

\begin{enumerate}
    \item We create a novel multilingual test dataset with quality-checked translations and new evaluation metrics.
    \item We introduce a scalable projection technique by integrating the LASER multilingual encoder with popular open-source LLMs like CodeLLaMa \cite{roziere2023code}, CodeGemma \cite{team2024codegemma}, and Mistral \cite{jiang2023mistral} for zero-shot cross-lingual code generation.
    \item Our evaluation of the approach against Chain-of-Thought (CoT) and fine-tuning with multilingual bootstrapped data, highlights the strengths and limitations of each method.
\end{enumerate}

We make our code\footnote{\url{https://github.com/lmd0420/Multilingual_Code_Gen}} and multilingual evaluation data\footnote{\url{https://huggingface.co/datasets/AnonymousDoe/MBPP-Translated}} publicly available for academic use.

\section{Related Work}
As AI technology advances, transformer-based LLMs like GPT \cite{openai2023gpt4}, LLaMA \cite{touvron2023llama}, Mistral \cite{jiang2023mistral}, and Gemma \cite{team2024gemma} have become prominent in research and applications. While pre-trained models such as LLaMA2 and Gemma are fine-tuned for code generation, their English-centric training data limits multilingual proficiency \cite{lai2023chatgpt, akiki2022bigscience}. Studies show these models face performance issues with non-English tasks \cite{shi2022language, becker2023programming}, and human supervision remains crucial for quality \cite{sarsa2022automatic}.

To address these gaps, Ahuja et al. \cite{ahuja2023mega} develop a multilingual benchmark for evaluating LLMs, revealing performance drops across languages. Tan et al. \cite{tan2020evaluating} and Huang et al. \cite{huang2023languages} suggest leveraging transfer learning and cross-lingual prompting to improve multilingual capabilities. Additional methods include language-specific pre-training \cite{pfeiffer-etal-2022-lifting} and consistency regularization for fine-tuning \cite{zheng-etal-2021-consistency}.

Optimizing prompts enhances multilingual LLM accuracy \cite{zhao2021discrete, huang2022zero}, and CoT techniques improve code generation \cite{ma2023bridging}. Fine-tuning with multilingual synthetic data, including translation and back-translation \cite{sennrich2015improving, hoang2018iterative}, further refines LLMs \cite{li2023self, chai2023erniecodeenglishcentriccrosslingualpretraining, workshop2023bloom176bparameteropenaccessmultilingual, nakamura2024auroramopensourcemultilingual, zhang2024gettinglesslargelanguage}. Contrary to these popular approaches, we take an orthogonal route by using specialized multilingual encoders and lightweight projectors to bridge language gaps in popular LLMs. Our work is inspired by multimodal AI literature, integrating projection techniques for different modalities such as language, vision and speech \cite{liu2024visual,fathullah2024prompting,beyer2024paligemma}.

\section{Experimental Setup}
\label{ref:assess}
This section details our experimental setup, including the creation of a multilingual benchmark dataset, the models evaluated, and the metrics used. Our focus is on Python code generation from multilingual prompts, though the methods and insights are applicable to other languages and contexts.
\subsection{Evaluation Dataset}
\label{ref:evaldata}
To obtain datasets with multilingual prompts for code generation, we adapted the Mostly Basic Programming Problems (MBPP) dataset \cite{austin2021program}, specifically the sanitized version with its "test" split, containing $257$ problems with solutions and three test cases each. We translate these prompts into five languages—Chinese-Simplified (\textit{zh-cn}), Spanish (\textit{es}), Japanese (\textit{ja}), Russian (\textit{ru}), and Hindi (\textit{hi})—using the Google Translate API\footnote{https://cloud.google.com/translate}. Our selection covers languages with similar characters to English (e.g., Spanish) as well as those with entirely different characters (e.g., Japanese), demonstrating robustness across different language families.

Translation quality was assessed by: 
\begin{itemize}
    \item (a) expert bilingual speakers via \textit{Amazon Mechanical Turk}, who rate translations as acceptable or not. (Note: guidelines were provided for binary rating and consent was obtained to report the statistics in the paper.)
    \item (b) GPT-4, which rate translations on a scale of 1 to 5.
\end{itemize}
Human-evaluated results in table \ref{tab:acceptability_agreement} show superior  translation quality. Table \ref{tab:trans_quality} presents GPT-4's ratings, again indicating that translations are of high-quality with high mean scores and low standard deviations.

\begin{table}[t]
\centering
\resizebox{\columnwidth}{!}{%
\begin{tabular}{lccc}
\toprule
\textbf{Translation} & \textbf{A1} & \textbf{A2} & \textbf{Agreement (\%)} \\
\midrule
en\_es & 0.94 & 0.96 & 89.69 \\
en\_hi & 0.93 & 0.96 & 89.11 \\
en\_ja & 0.93 & 0.96 & 89.88 \\
en\_ru & 0.93 & 0.96 & 90.43 \\
en\_zh-cn & 0.94 & 0.96 & 90.79 \\
\bottomrule
\end{tabular}
}
\caption{Human Evaluation of Translated Prompts. Two distinct bilingual speakers from MTurk rated translations with 1 (acceptable) or 0 (not acceptable) for each translation. A1 and A2 represent their average scores.}
\label{tab:acceptability_agreement}
\end{table}

\begin{table}[t]
    \centering
    \resizebox{\columnwidth}{!}{%
    \begin{tabular}{l c c}
        \toprule
        \textbf{Lang. Pair} & \textbf{Average Rating} & \textbf{St.Dev} \\
        \midrule
        en\_hi & 4.88 & 0.40 \\
        en\_es & 4.90 & 0.48 \\
        en\_ru & 4.95 & 0.30 \\
        en\_zh-cn & 4.93 & 0.39 \\
        en\_ja & 4.87 & 0.55 \\
        \bottomrule
    \end{tabular}
    }
    \caption{Average Rating and Standard Deviation for Translation from English to Other Languages}
    \label{tab:trans_quality}
\end{table}

\begin{figure*}[t]
    \centering
    \begin{minipage}[b]{\textwidth}
        \centering
        \includegraphics[width=\textwidth]{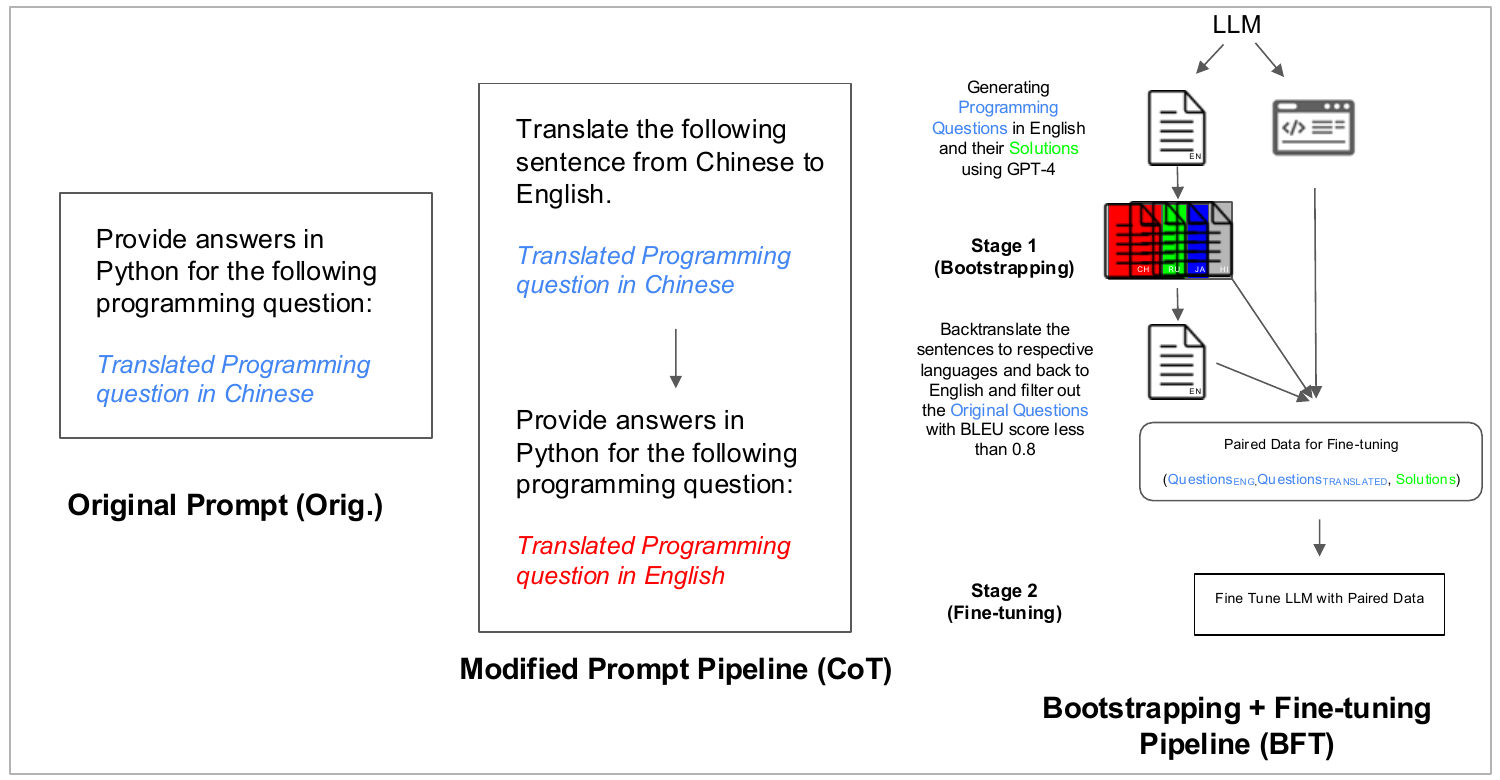} % Replace with your image
        \subcaption{Baselines with direct prompting, Chain of Thought (CoT) and fine-tuning with bootstrapped data}
        \label{fig:subfig2}
    \end{minipage}
    \vfill
    \begin{minipage}[b]{\textwidth}
        \centering
        \includegraphics[width=\textwidth]{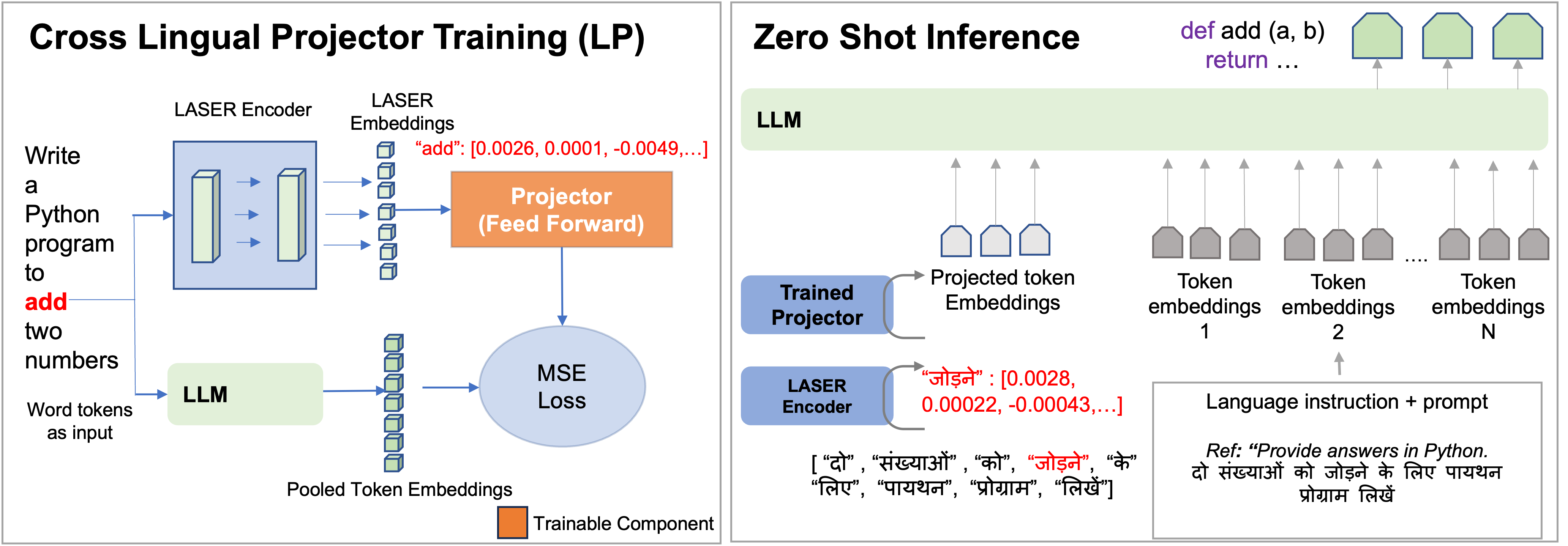} % Replace with your image
        \subcaption{Our proposed approach based on cross lingual encoder and projector training and zero shot inference}
        \label{fig:subfig1}
    \end{minipage}
    \caption{Explored approaches}
    \label{fig:explorations}
\end{figure*}

\subsection{Models Used for Evaluation}
We consider three open source variants of instruction-tuned models for evaluation, namely \textsc{CodeLLaMa-7B}\footnote{codellama/CodeLlama-7b-Instruct-hf}, \textsc{CodeGemma-7B}\footnote{google/codegemma-7b-it} and \textsc{Mistral-7B-v0.3}\footnote{mistralai/Mistral-7B-Instruct-v0.3}. These models are specialized versions of their base models to programming-related tasks. They have demonstrated greater efficacy at code generation, infilling, and debugging capabilities compared to the standard versions. Models are obtained from HuggingFace hub\footnote{\url{http://huggingface.co}}. For benchmarking, we adopt GPT-4 as the reference system (\textit{a.k.a Skyline}) because it has been extensively trained on various languages, and has proven effective in various tasks, including code generation. 

\subsection{Inference Pipeline and Evaluation Metrics}
We develop a pipeline to process task descriptions in various languages. The pipeline feeds these prompts into the models and variants described in Sections \ref{sec:baselines} and \ref{sec:projection} and stores the results. We aim to provide a solution for people with little programming experience who use a variety of languages, allowing them to access LLMs’ coding capabilities through natural language. Thus, we choose to report our results on from-scratch code generation tasks in the zero-shot setting. Python code from the outputs is extracted using regular expressions. We then automatically identify function names in the code, replacing the function names in the MBPP test assertions with those from the model outputs, ensuring any function names would not affect the test results. Finally, we generate bash scripts to execute the extracted code and assertions, and measure the following metrics:

\begin{itemize}
    \item \textbf{Total Error Rate (TotalER):} The ratio of code samples that fail at least one test case, to the total number of samples. Lower is better.
    \item \textbf{Logical Error Rate (LER):} The ratio of code samples that execute successfully but produce incorrect results, to the total number of samples. Lower is better.
    \item \textbf{Syntax Error Rate (SER):} The ratio of code samples containing syntax errors, to the total number of samples. Lower is better.
    \item \textbf{All Tests Passed Rate (ATPR):} The ratio of code samples that pass all given test cases, to the total number of samples. Higher is better.
\end{itemize}

Additionally, we also observe the \textit{Code Completion Rate} as a supplementary metric, which indicates the proportion of complete codes in model responses. A higher value represents a better result.

With this setup, we can now evaluate LLM code generation quality and propose mitigation strategies. Our approach and baselines are summarized in Figure \ref{fig:explorations}, detailed in the following section.

\begin{table*}[t]
    \centering
    \setlength{\tabcolsep}{4pt} % Adjust the space between columns
    \resizebox{\textwidth}{!}{
    \begin{tabular}{llcccc|cccc|cccc|cccc}
        \toprule
        \multirow{2}{*}{\textbf{LLM}} & \multirow{2}{*}{\small{Lang}} & \multicolumn{4}{c}{\textbf{TotalER}\textcolor{BrickRed}{\textbf{$\downarrow$}}} & \multicolumn{4}{c}{\textbf{LER} \textcolor{BrickRed}{\textbf{$\downarrow$}}} & \multicolumn{4}{c}{\textbf{SER} \textcolor{BrickRed}{\textbf{$\downarrow$}}} & \multicolumn{4}{c}{\textbf{ATPR}\textcolor{Green}{\textbf{$\uparrow$}}} \\
        \cmidrule(lr){3-6} \cmidrule(lr){7-10} \cmidrule(lr){11-14} \cmidrule(lr){15-18}
        & & \textbf{Orig.} & \textbf{CoT} & \textbf{BFT} & \textbf{LP} & \textbf{Orig.} & \textbf{CoT} & \textbf{BFT} & \textbf{LP} & \textbf{Orig.} & \textbf{CoT} & \textbf{BFT} & \textbf{LP} & \textbf{Orig.} & \textbf{CoT} & \textbf{BFT} & \textbf{LP} \\
        \midrule
        \multirow{6}{*}{\makecell{GPT-4 \\(Skyline)}} & en & \textit{58.37} & - & - & - & \textit{10.9} & - & - & - & \textit{47.47} & - & - & - & 41.63 & - & - & - \\
        & es & \textit{62.65} & - & - & - & \textit{12.85} & - & - & - & \textit{49.8} & - & - & - & \textit{37.35} & - & - & - \\
        & hi & \textit{67.7} & - & - & - & \textit{17.9} & - & - & - & \textit{49.8} & - & - & - & \textit{32.3} & - & - & - \\
        & ja & \textit{64.2} & - & - & - & \textit{13.62} & - & - & - & \textit{50.58} & - & - & - & \textit{35.8} & - & - & - \\
        & ru & \textit{65.37} & - & - & - & \textit{17.12} & - & - & - & \textit{48.25} & - & - & - & \textit{34.63} & - & - & - \\
        & zh & \textit{67.7} & - & - & - & \textit{16.73} & - & - & - & \textit{50.97} & - & - & - & \textit{32.3} & - & - & - \\
        \midrule
        % \multirow{6}{*}{GPT-3.5} & en & 100.0 & - & 100.0 & - & 21.01 & - & 17.12 & - & 51.75 & - & 50.58 & - & 27.24 & - & 32.3 & - \\
        % & es & 100.0 & 100.0 & 99.61 & - & 20.23 & 21.79 & 16.73 & - & 53.7 & 51.75 & 52.14 & - & 26.07 & 26.46 & 31.13 & - \\
        % & hi & 100.0 & 99.61 & 100.0 & - & 29.19 & 27.63 & 23.34 & - & 51.36 & 52.14 & 54.09 & - & 19.45 & 20.23 & 22.57 & - \\
        % & ja & 100.0 & 100.0 & 100.0 & - & 23.74 & 26.84 & 20.63 & - & 50.58 & 50.2 & 50.97 & - & 25.68 & 22.96 & 28.4 & - \\
        % & ru & 99.61 & 99.61 & 100.0 & - & 20.62 & 23.74 & 20.23 & - & 51.36 & 49.8 & 52.53 & - & 28.02 & 26.46 & 27.24 & - \\
        % & zh & 100.0 & 100.0 & 100.0 & - & 20.23 & 22.95 & 16.73 & - & 52.53 & 50.2 & 51.75 & - & 27.24 & 26.85 & 31.52 & - \\
        % \hline
        \multirow{6}{*}{\makecell{Code\\LLaMa-7B}} & en & 87.16 & - & 82.1 & \textbf{75.49} & 63.04 & - & 28.79 & \textbf{22.57} & \textbf{24.12}$^*$ & - & 53.31 & 52.92 & 12.84 & - & 17.9 & \textbf{24.51} \\
        & es & 79.77 & 91.83 & \textbf{81.71} & \textbf{81.71} & 28.8 & 56.81 & 26.07 & \textbf{24.9} & 50.97 & \textbf{35.02}$^*$ & 55.64 & 56.81 & \textbf{20.23} & 8.17 & 18.29 & 18.29 \\
        & hi & 96.5 & 97.66 & 96.5 & \textbf{95.72} & 65.37 & 61.08 & 61.87 & \textbf{25.29} & \textbf{31.13} & 36.58 & 34.63 & 70.43 & 3.5 & 2.34 & 3.5 & \textbf{4.28} \\
        & ja & 89.49 & 84.82 & 84.82 & \textbf{84.44} & 50.58 & 52.91 & 34.24 & \textbf{22.96} & 38.91 & \textbf{31.91} & 50.58 & 61.48 & 10.51 & 15.18 & 15.18 & \textbf{15.56} \\
        & ru & \textbf{82.1} & 86.38 & 85.21 & 82.88 & 39.69 & 61.87 & 31.51 & \textbf{23.35} & 42.41 & \textbf{24.51} & 53.7 & 59.53 & 17.9 & 13.62 & 14.79 & \textbf{17.12} \\
        & zh & 93.77 & 96.5 & 88.72 & \textbf{82.1} & 77.43 & 73.15 & 35.41 & \textbf{26.46} & \textbf{16.34} & 23.35 & 53.31 & 55.64 & 6.23 & 3.5 & 11.28 & \textbf{17.9} \\
        \midrule
        \multirow{6}{*}{\makecell{Code\\Gemma-7B}} & en & 82.1 & - & 92.22 & \textbf{77.04} & 41.63 & - & 63.04 & \textbf{25.68} & 40.47 & - & \textbf{29.18} & 51.36 & 17.9 & - & 7.78 & \textbf{22.96} \\
        & es & 86.38 & 89.1 & 91.05 & \textbf{77.82} & 47.86 & 42.02 & 57.59 & \textbf{24.51} & 38.52 & 47.08 & \textbf{33.46} & 53.31 & 13.62 & 10.9 & 8.95 & \textbf{22.18} \\
        & hi & 89.49 & 91.05 & 94.16 & \textbf{81.71} & 49.41 & 50.58 & 74.71 & \textbf{29.18} & 40.08 & 40.47 & \textbf{19.45} & 52.53 & 10.51 & 8.95 & 5.84 & \textbf{18.29} \\
        & ja & 83.66 & 90.27 & 91.05 & \textbf{79.77} & 38.91 & 44.75 & 50.58 & \textbf{24.13} & 44.75 & 45.52 & \textbf{40.47} & 55.64 & 16.34 & 9.73 & 8.95 & \textbf{20.23} \\
        & ru & 85.99 & 88.72 & 89.1 & \textbf{77.04} & 42.41 & 48.25 & 59.53 & \textbf{25.68} & 43.58 & 40.47 & \textbf{29.57} & 51.36 & 14.01 & 11.28 & 10.9 & \textbf{22.96} \\
        & zh & 84.82 & 86.38 & 93.0 & \textbf{79.38} & 39.68 & 48.64 & 62.26 & \textbf{28.02} & 45.14 & 37.74 & \textbf{30.74} & 51.36 & 15.18 & 13.62 & 7.0 & \textbf{20.62} \\
        \midrule
        \multirow{6}{*}{\makecell{Mistral\\-7B-v0.3}} & en & 85.21 & - & 92.61 & \textbf{83.27} & 35.41 & - & 28.41 &\textbf{27.24} & \textbf{49.8} & - & 64.2 & 56.03 & 14.79 & - & 7.39 & \textbf{16.73} \\
        & es & 87.55 & 86.38 & 94.94 & \textbf{84.82} & 39.69 & 29.18 & 26.46 & \textbf{26.06} & \textbf{47.86} & 57.2 & 68.48 & 58.76 & 12.45 & 13.62 & 5.06 & \textbf{15.18} \\
        & hi & 91.44 & \textbf{91.05} & 98.83 & 92.22 & 35.41 & 35.41 & \textbf{24.12} & 30.74 & 56.03 & \textbf{55.64} & 74.71 & 61.48 & 8.56 & \textbf{8.95} & 1.17 & 7.78 \\
        & ja & 88.72 & \textbf{86.77} & 96.11 & 87.55 & 35.8 & 31.91 & 28.02 & \textbf{22.57} & \textbf{52.92} & 54.86 & 68.09 & 64.98 & 11.28 & \textbf{13.23} & 3.89 & 12.45 \\
        & ru & 85.6 & \textbf{84.05} & 95.33 & \textbf{84.05} & 33.85 & 30.74 & 26.85 & \textbf{24.13} & \textbf{51.75} & 53.31 & 68.48 & 59.92 & 14.4 & \textbf{15.95} & 4.67 & \textbf{15.95} \\
        & zh & 88.72 & 87.16 & 94.55 & \textbf{84.05} & 39.3 & 30.74 & \textbf{26.07} & 26.85 & \textbf{49.42} & 56.42 & 68.48 & 57.2 & 11.28 & 12.84 & 5.45 & \textbf{15.95} \\
        
        % Add additional rows for each model and language with their respective data points
        \bottomrule
    \end{tabular}}
    \caption{Comprehensive comparison of different models across multiple languages and configurations. TotalER: Total Error Rate, LER: Logical Error Rate, SER: Syntax Error Rate, ATPR: All Test Passed Rate. \textit{Orig:} Directly Querying LLMs, \textit{CoT}: Chain of Thought with Translation, \textit{BFT}: Fine tuning on Bootstrapped Multilingual Data, \textit{LP} (Our approach): Fine tuning on Multilingual Projection with LASER Encoders}
    \label{tab:overall_results}
\end{table*}

\section{Issues with Trivial Baselines}
\label{sec:baselines}
Given that language models exhibit emergent capabilities and scale effectively across tasks and languages, efficient prompting and prompt tuning are generally preferred over costly training or fine-tuning that demands extensive data curation. Based on our experimental setup, we highlight the challenges LLMs face with multilingual code generation in conventional settings, providing a detailed analysis of existing models' performance and their limitations. Throughout this section, we will reference Table \ref{tab:overall_results} for a comprehensive discussion of the results.

\subsection{Baseline 1. Original Prompt}
Here, each query in the dataset is passed through the pipeline, where the model generates response code, filtered from extraneous information such as code explanations, and executed using an automatically constructed bash script. The results are presented in first column of each section of Table \ref{tab:overall_results}, with the following key observations:

GPT-4, recognized for its robustness and extensive engineering, reliably generates code across all language prompts, though with slightly varying error profiles - except for Hindi and Chinese. In contrast, open-source models like CodeLLaMa show more pronounced disparities between languages, with higher error rates and lower all-tests-passed rates compared to English. Notably, some models, such as CodeLLaMa-Instruct-7B, perform better in non-English languages like Spanish. This may seem unusual but aligns with findings from Chen et al.\cite{chen2024monolingualmultilingualinstructiontuning}, which show that LLaMa 7B, when instruction-tuned for multilingual tasks, performs better in Spanish than English. Since CodeLLaMa is based on this instruction-tuned model, this could explain the atypical performance in Spanish. Overall, these results highlight a lack of consistency in code output quality as the language changes. We use the abbreviation \textit{Orig.} to refer to this baseline henceforth.

\subsection{Chain-of-Thought with Back-Translation}
Due to uneven language representation in LLM training datasets, achieving consistent results with direct prompting is challenging. A potential solution is to use back-translation: translate non-English prompts into English and use the English version as the query. This CoT approach involves translating the problem statement with the prompt: \texttt{Translate the sentence \$PROBLEM from \$TARGET-LANG to English}, then generating code outputs from the translated prompt. Our experiments, detailed in the second column of Table \ref{tab:overall_results}, show that back-translation does not significantly improve results. In some cases, it even reduce performance, as indicated by lower ATPR scores. Qualitative analysis suggests that models struggle with non-canonical language representations and topic drift, despite the translations not being of poor quality. We use the abbreviation \textit{CoT} to refer to this baseline henceforth.

\subsection{Bootstrapping Multilingual Data and Fine Tuning}
Fine-tuning pre-trained models is effective for many NLP tasks but is often resource-intensive, requiring costly and time-consuming task-specific labeled data. Instead of manually creating such data for multiple languages - designing prompts, validating answers, and translating while preserving semantic meaning - we use a bootstrapping approach. In this method, we utilize a powerful LLM like ChatGPT to generate English programming problems and their answers. These problems are then translated into target languages and back-translated into English. We assess the similarity of translations using the BLEU score \cite{papineni2002bleu}, retaining translations that meet a quality threshold (e.g., 0.8) to create new training data. This method preserves text quality in target languages and allows the model to validate its translations, as detailed in Algorithm \ref{alg:bootstrap}.

\begin{algorithm}

\caption{Bootstrap Training Data}
\label{alg:bootstrap}
\begin{algorithmic}[1]
\Function{BootstrapData}{$LLM,$Lang}
  \State $n \gets$ number of attempts
  \State $threshold \gets$ 0.9
  \State Initialize a query set $Q \gets$ \{\}
  \State Initialize training data $TD \gets$ \{\}
  \State $squery \gets$ ''Generate 100 python problems''
  \State $trquery \gets$ ''Translate from English into $Lang$''
  \State $btrquery \gets$ ''Translate from $Lang$ into English''
  \For{$i \gets 1$ to $n$}
    \State  $q \gets$ \Call{LLM}{$squery$}
    \State Push $q$ into query set $Q$
  \EndFor
  
  \For {$q$ in $Q$}
    \State  $a \gets$ \Call{LLM}{$q$}
    \State Push $a$ into answer set $A$
    \State Push $<q,a>$ into $TD$
  \EndFor
  
  \For {$q,a$ in $TD$}
    \State  $t \gets$ \Call{LLM}{$trquery$, $q$}
    \State  $bt \gets$ \Call{LLM}{$btrquery$, $t$}
    \State $score \gets$ \Call{BLEU}{$t$,$bt$}
    \If {$score > threshold$}
      \State Push $<t,a>$ into $TD$
    \EndIf
  \EndFor
  \State \Return $TD$
\EndFunction
\end{algorithmic}
\end{algorithm}

After bootstrapping data for all target languages, we shuffle and use it to fine-tune the LLMs with a single A100 GPU. Models are quantized to FP16 and fine-tuned using parameter-efficient techniques, including low-rank adaptation \cite{hu2021lora}. We set the temperature to 0.8 for consistency and use two epochs. As shown in the third column of Table \ref{tab:overall_results}, while bootstrapping with ChatGPT reduces syntax errors, it also increases hallucinations, leading to lower test pass rates and higher total errors. This suggests that the model, although producing more complete code, struggles with accuracy and reliability. We use the abbreviation \textit{BFT} to refer to this baseline henceforth.

\section{Our Approach: Projection-Based Zero-Shot Transfer} 
\label{sec:projection}
Our approach focuses on avoiding the use of in-language training data, which can be costly and impractical. Instead, we utilize an intermediate, lightweight method that relies on abundant English data and the LASER multilingual encoder \cite{artetxe2019massively}, which provides joint embeddings for over 100 languages. In this setup, the LASER encoder preprocesses and embeds input tokens before passing them to the LLM, which then operates on these embeddings rather than raw input IDs. This method enables efficient language scaling, as similar meanings are represented consistently across languages (e.g., the English token "add" and its Hindi counterpart "JoDaNe" are embedded similarly, as shown in Figure \ref{fig:explorations} part (b)).

Two key challenges arise with this approach: (A) differing tokenization between the multilingual encoder and the LLM, and (B) the LLM's unfamiliarity with the multilingual embeddings. To address (A), we use word tokens and extract mean-pooled embeddings from subwords using tokenizers such as NLTK \footnote{https://www.nltk.org} for space-sparated lanaguge inputs, Jieba \footnote{https://github.com/fxsjy/jieba} for Chinese, and Janome\footnote{https://mocobeta.github.io/janome/en/} for Japanese. We then train a projector to align these embeddings. For a given word token, we compute the LLM’s subword embeddings (\(\hat{H}_{llm}\)) through max pooling, and the multilingual embeddings (\(H_{laser}\)) from the LASER encoder. The projector, with learnable parameters \(\mathbf{W}_{llm}\) and \(\mathbf{b}_{llm}\), is defined as:

\[
\mathbf{H}_{llm} = \mathbf{W}_{llm} \cdot \mathbf{H}_{laser} + \mathbf{b}_{llm}
\]

The model is trained by minimizing the Mean Squared Error (MSE) between \(\hat{H}_{llm}\) and \(\mathbf{H}_{llm}\):

\[
\text{MSE} = \frac{1}{N} \sum_{i=1}^{N} \left\| \hat{H}_{llm}^i - \mathbf{H}_{llm}^i \right\|^2
\]

where \(N\) is the number of word tokens. Training utilizes English tokens from the MBPP dataset, which includes 120 examples. We train the projector for 200 epochs on a single consumer‑grade NVIDIA RTX-4060; the entire run finishes in under one hour. In contrast, LoRA fine‑tuning or re‑pretraining on extra multilingual data typically requires at least several GPU hours or even days, highlighting an orders‑of‑magnitude efficiency advantage for our lightweight projector training.

During inference, tokens are first word-tokenized and embedded using LASER, then projected, and finally input to the LLM for multilingual processing without requiring in-language data. To enhance performance and align with baselines, we also concatenate system prompt embeddings with the original programming prompt embeddings. Notably, LASER embeddings are of size 1024, while LLM embeddings are typically 4096 or larger, necessitating a 4-fold upsampling. We achieve this using two linear projection layers as outlined in the above equations. We use the abbreviation \textit{LP} to explain this system henceforth.

\section{Results and Discussions}
\label{sec:expres}

\begin{figure*}[t]
  \centering
  \includegraphics[width=14.5cm]{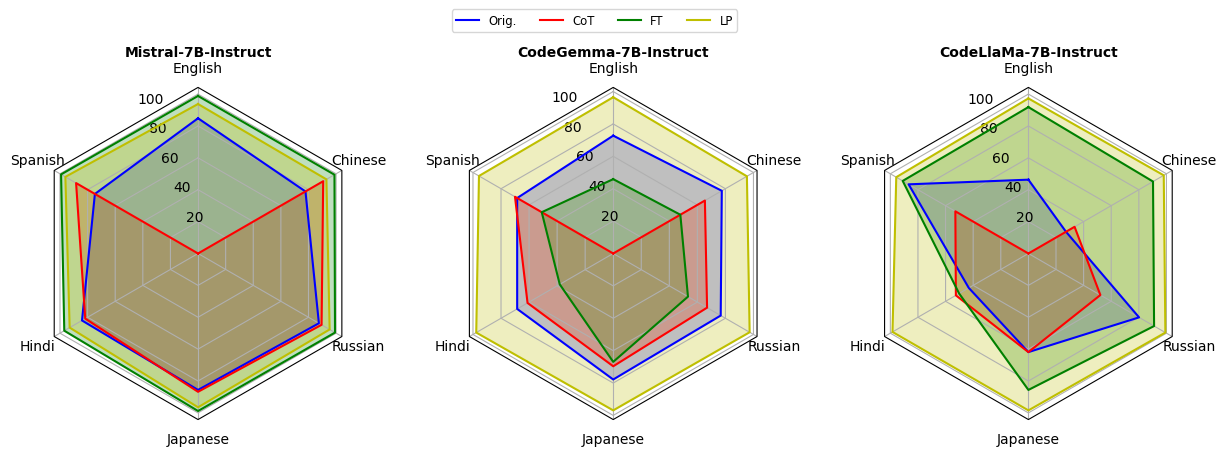}
  \caption{Code Completion Rate (CCR) for Models and Languages, with LP represented by perfect polygons, always above 90\%}
  \label{fig:CCR}
\end{figure*}

Table \ref{tab:overall_results} presents the overall performance models and variants discussed in sections \ref{sec:baselines} and \ref{sec:projection}.  Our observations indicate that across all metrics, our proposed model consistently reduces the performance gap between English and non-English languages, as reflected in the differences and deviations. This improvement is particularly evident when comparing the direct querying setup (Orig.) with our multilingual projector-based variant (LP), where deviations from English are generally smaller. It is also worth noting that the projection method we proposed completely eliminates the need for external data. Therefore, we compared it with methods like bootstrapping from the model itself followed by finetuning (BFT) and Chain-of-Thought (CoT), which also do not rely on external data or models. We explore the details of each metric below.

\subsection{Total Error Rate (TotalER)}
TotalER is an important metric that quantifies the overall error rate of the generated code. Our proposed method, LP, consistently achieves the lowest TotalER across nearly all languages and models, demonstrating its effectiveness. For example, with the CodeLLaMa-7B model, LP significantly reduces the TotalER to $75.49$ for English (en) and $82.1$ for Chinese (zh), outperforming the Orig. and other methods. This improvement is especially pronounced in languages with complex syntax and morphology, such as Hindi (hi) and Russian (ru), where LP reduces the TotalER by over 10\% in some cases compared to the original model. Even in cases where LP is the second-best, its performance is very close to the top-performing method, highlighting its reliability. In contrast, BFT, a strong trivial baseline, tends to increase the TotalER due to hallucinations, as observed in our data analysis, despite slightly improving the all test cases passed metric.

\subsection{Logical Error Rate (LER)}
LER is a critical component of the total error, measuring the proportion of code samples that execute without errors but produce incorrect results. A lower LER indicates a model’s ability to generate logically sound code, making it a key metric for evaluating performance. It’s important to note that we classify a logical error not only when no valid code is generated but also when any of the test cases fail.

Our approach, LP, consistently outperforms other methods in terms of LER, with only a few exceptions where the difference is marginal and still better than other candidates. For instance, with the CodeGemma-7B model, LP achieved an LER of 25.68 for English, significantly lower than the 41.63 in Orig and 63.04 in BFT. This trend is also evident in other languages, such as Spanish (es) and Japanese (ja), where LP substantially reduces LER, underscoring its effectiveness in ensuring logical correctness across multilingual scenarios.

\subsection{Syntax Error Rate (SER)}

SER is a component of total error and indicates the proportion of code samples that contain syntax errors. A lower SER reflects the model's ability to generate syntactically correct code. Our overall observations with respect to this metric is models like ours that often produce code than omitting it (which is indicated by the higher code completion rate) are more prone to syntax error due to the high recall. While syntax error solving is a crucial step in program debugging, we believe such a form of error is slightly easier for beginners to solve than logical errors. Thus, given that LP consistently achieves the lowest LER across all languages and models, we consider this demonstrates the LP's proficiency in helping with generating error-free code, particularly in linguistically diverse contexts.

\subsection{All Test Passed Rate (ATPR)}

ATPR measures the proportion of code samples that successfully pass all given test cases. A higher ATPR signifies greater reliability of the generated code, making it a crucial metric. Our observations show that LP consistently outperforms other methods in terms of ATPR across most cases. Also, the improvement in English performance may be due to LASER slightly altering the way text enters the model. However, there are exceptions with the Mistral-7B-v0.3 model in a few languages. This model, being more recent, benefits from enhanced multilingual capabilities due to its diverse pretraining datasets and extended vocabulary. Overall, ATPR improvements are consistent across other languages, highlighting LP's superior performance in generating reliable and functional code.

Our observations using Multilingual Projections with LASER encoders reveal that LP not only reduces errors but also enhances the logical correctness and reliability of the generated code, establishing it as the leading approach for multilingual Python code generation. Additionally, we analyze the Code Completion Rate (CCR) to assess the robustness of these models in generating meaningful code rather than nonsensical explanations across languages. LP consistently outperforms other variants in this regard, as shown in the spider graph in Figure \ref{fig:CCR}. This graph illustrates LP’s strong performance in producing complete code across all languages. Notably, the shapes representing LP in the graph are perfect polygons, reflecting its consistent behavior and reliability across different languages.

\section{Conclusions and Future Work}
\label{sec:concl}
In this paper, we demonstrate the significant potential of Large Language Models to bridge language gaps and promote inclusivity in multilingual prompt-based code generation. While LLMs exhibit promising capabilities across various languages, their performance can be inconsistent, particularly with non-English prompts. Our comprehensive analysis and evaluation using a revised benchmark dataset revealed both strengths and limitations in multilingual code generation, highlighting areas needing improvement.

We showcase the limited effectiveness of traditional Chain-of-Thought method, as well as bootstrapping multilingual training data and fine-tuning LLMs to enhance code generation quality across multiple languages. Our zero-shot cross-lingual transfer approach, utilizing projected embeddings, proves effective, as evidenced by improved ATPR and reduced TotalER values. A single-layer projector already narrows most of the cross‑lingual gap to within a 2‑percentage‑point range; this suggests that more complex projector architectures are unnecessary, as their potential accuracy gains are marginal while their memory and compute overheads would grow substantially.

Our lightweight method eliminates the need for extensive external multilingual data, maximizing the model's potential internally, and can be easily extended to other multilingual encoder + LLM combinations. Because the projector operates purely in embedding space, exactly the same training recipe can transfer seamlessly from the original LSTM‑based LASER encoder to the newer transformer‑based LASER3\cite{heffernan2022bitextminingusingdistilled} variant, underscoring the generality of our approach.

Future work will expand this approach to include more languages, diverse prompt patterns, and programming languages beyond Python. Our findings underscore the importance of advancing these techniques to enhance LLM adaptability and utility for a global audience, stressing the need for ongoing efforts to improve their effectiveness and versatility in diverse linguistic contexts.

\section{Limitations}
A major limitation of this work lies in the reliance on word tokenization and pooled token embeddings, which introduces external dependencies and may not scale effectively to extremely low-resource languages where tokenizers are not readily available. Furthermore, the sequence of projected embeddings from the target language can significantly differ from the canonical English order, potentially hindering the model’s ability to fully leverage these embeddings. This misalignment could contribute to the generation of hallucinatory and erroneous outputs. To address this issue, some degree of fine-tuning of LLMs with denoising objectives may be necessary. 

Additionally, our study focuses solely on generating code from scratch and does not cover code-filling scenarios, which is another important aspect that warrants future exploration. Due to resource constraints, the scope of this study has been limited to Python, but expanding the approach to encompass other general-purpose and special-purpose programming languages, as well as more benchmarks \cite{wang2023executionbasedevaluationopendomaincode, peng2024humanevalxlmultilingualcodegeneration} is essential for broader applicability.

\section*{Acknowledgment}
The models we utilized in our study are widely used ones from OpenAI, Google, MistralAI and Meta, and we employed Google Cloud Translator and the MBPP dataset on Hugging Face. All of these resources are publicly accessible; we did not introduce any additional real-world data, thus avoiding the creation of new ethical and privacy issues.

Given we are dealing with black-box Large Language Models as part of this study, there needs to be careful consideration of any potential biases that can be harmful in nature. Although we are focusing on a objective task with little to no opinion sourcing from the models, cultural and racial biases can occur given we are exposing the models to multi-lingual prompts. Since the applications we are focusing on are essentially user-centric in nature, a proper communication protocol should be established that can help clarify potential erratic behaviour of models, especially for low-resource languages. We would also like to share that we employed OpenAI's ChatGPT-4 system to enhance writing efficiency by generating LaTeX code, ensuring concise sentences, and aiding in error debugging.

\bibliographystyle{IEEEtran} 
\bibliography{references}

\end{document}